\documentclass[12pt]{article}
\usepackage{mathptmx}
\usepackage{amsmath,amssymb}
\usepackage{natbib}
\usepackage{dsfont}
\usepackage{xcolor}
\usepackage{gensymb}
\usepackage{rotating}
\usepackage{subcaption}
\usepackage{booktabs}
\usepackage{tabularx}
\usepackage{array}
\usepackage{placeins}
\usepackage{multirow,float,needspace}
\usepackage[hidelinks]{hyperref}
\usepackage[shortlabels]{enumitem}
\usepackage{graphicx} \usepackage{tikz}
\usetikzlibrary{arrows.meta,positioning}
\usepackage[margin=1in]{geometry}
\usepackage{amsthm}

\usepackage{libertine}
\usepackage{microtype} %
\usepackage[most]{tcolorbox} %
\usepackage{fontawesome5} %
\definecolor{columbiablue}{HTML}{B9D9EB} %
\usepackage{cleveref}
\usepackage[all,defaultlines=3]{nowidow}

\crefname{section}{Section}{Sections}
\Crefname{section}{Section}{Sections}
\crefname{subsection}{Section}{Sections}
\Crefname{subsection}{Section}{Sections}
\crefname{equation}{Equation}{Equations}
\Crefname{equation}{Equation}{Equations}
\crefname{figure}{Figure}{Figures}
\Crefname{figure}{Figure}{Figures}
\crefname{table}{Table}{Tables}
\Crefname{table}{Table}{Tables}
\crefname{appendix}{Appendix}{Appendices}
\Crefname{appendix}{Appendix}{Appendices}

\newcommand{\CE}{\operatorname{CE}}
\newcommand{\KL}{\operatorname{KL}}
\newcommand{\sg}{\operatorname{sg}}
\newcommand{\softmax}{\operatorname{softmax}}
\newcommand{\appendixfigurebreak}{\FloatBarrier}
\newcommand{\interactivefigures}{Interactive versions of both figures, with animations that interpolate between input and output positions, are available at {\urlstyle{same}\url{https://jingjingz.github.io/WearableBridge/}}.}

\makeatletter
\g@addto@macro\appendix{%
  \crefalias{section}{appendix}%
  \crefalias{subsection}{appendix}%
  \begin{center}
    \Large{\textbf{Appendix}}
  \end{center}}
\makeatother

\title{WearableBridge: Task-Guided Latent Transport\\for Free-Living Activity Recognition}

\author{
  Jingjing Zou\thanks{Correspondence to: \href{mailto:j2zou@ucsd.edu}{j2zou@ucsd.edu}.}\\
  Herbert Wertheim School of Public Health\\
  and Human Longevity Science\\
  University of California San Diego
  \and
  Yixin Wang\\
  Department of Statistics\\
  University of Michigan, Ann Arbor
  }

\usepackage{etoolbox}
\makeatletter
\patchcmd{\@maketitle}{\large\lineskip}{\normalsize\lineskip}{}{\PackageError{main-arxiv}{author-size patch failed}{}}
\makeatother

\date{}

\hypersetup{pdftitle={WearableBridge: Task-Guided Latent Transport for Free-Living Activity Recognition},pdfauthor={Jingjing Zou and Yixin Wang}}

\begin{document}

\maketitle

\begin{tcolorbox}[enhanced, colback=columbiablue!50, colframe=columbiablue!50, boxrule=0pt, arc=2mm,
  left=6mm, right=6mm, top=5mm, bottom=5mm, before skip=6pt, after skip=10pt]
\small
\noindent\textbf{Abstract}\par
\noindent
Wrist-worn accelerometers make activity monitoring feasible at scale, but incidental arm motion, device orientation, and shifting context blur the signal that separates activities. Calibration studies supply higher-quality references such as concurrent thigh recordings or supervised laboratory wrist sessions, but they are expensive, cover few participants, and are rarely available at deployment. How can limited reference data improve recognition for participants who wear only a wrist device? We propose WearableBridge, which holds the reference encoder and classifier fixed and learns a task-guided transport map, the bridge, that carries each wrist feature into the region the classifier associates with its activity. Reference features or the classifier's predictions indicate the destination, and a classification loss penalizes wrist features that land in the region of another activity. We instantiate WearableBridge with three bridges: Mapper regresses onto reference features, Distillation matches the classifier's class probabilities, and MeanFlow learns a one-step flow from wrist toward reference features. On 40 held-out free-living PAAWS participants, all three bridges improve macro-F1 in every training run, and the strongest, MeanFlow, raises mean macro-F1 from 35.1\% to 56.0\% with concurrent thigh references (Thigh2Wrist), and from 54.0\% to 63.5\% with group-matched laboratory wrist references (Lab2Freeliving). Fitted only on PAAWS, MeanFlow also raises macro-F1 on the CAPTURE-24 cohort from 42.8\% to 54.7\%. The gains replicate on selfBACK, hold across four reference encoders, and grow with temporal context. A study can therefore collect reference recordings for a calibration subset and improve wrist-only recognition for every other participant, training only a bridge of 0.13 to 2.44 million parameters while the reference encoder and classifier stay unchanged.

\par\medskip
\noindent\begin{tabular}{@{}c@{\hspace{0.5em}}l@{\hspace{1em}}l@{}}
\faGithub & \textbf{Code} & \href{https://github.com/jingjingz/WearableBridge}{github.com/jingjingz/WearableBridge}\\
\faDesktop & \textbf{Interactive demo} & \href{https://jingjingz.github.io/WearableBridge/}{jingjingz.github.io/WearableBridge}
\end{tabular}
\end{tcolorbox}

\section{Introduction}
Wrist-worn accelerometers record activity over many days with a single device and have made population studies of more than 100,000 participants feasible \citep{doherty2017}. The same placement makes the signal hard to interpret: a wrist recording mixes arm gestures, changes in device orientation, transitions, and daily context with the movement that distinguishes cycling, sitting, standing, and walking. These sources of variation are strongest in free-living (FL) recordings, collected as participants follow their daily routines without scripted activities or supervision. 
Classifiers trained under controlled conditions therefore lose accuracy on the FL recordings that population studies collect \citep{gyllensten2011,sasaki2016,pavey2017}.

Other recordings reveal activity more clearly. A thigh sensor gives a direct view of lower-body movement, and a supervised laboratory wrist session holds recording conditions steady and supplies observed labels. We call these \emph{higher-quality references}, meaning that they carry more information about activity. Both are expensive at scale: a thigh recording requires a second device and participants' compliance, a laboratory session requires staff time and visits, and FL labels can require prolonged observation or video review \citep{sasaki2016,potter2025}. A realistic study therefore collects references for a calibration subset and deploys wrist-only monitoring to everyone else. We ask: \emph{how can limited reference data improve wrist-only activity recognition for new participants who have no reference recording?}

Two calibration designs make this question concrete, each named by the direction in which activity information transfers. In Thigh2Wrist, a wrist window and a thigh window record the same movement at the same time, so each wrist window has its own reference, an instance-level target. In Lab2Freeliving, the wrist windows come from simulated free-living (SimFL) sessions, which imitate everyday conditions, and the references are laboratory (Lab) wrist windows from a separate session. Matching by participant, activity, wrist side, and device placement gives a group-level target. In both designs, deployment uses only a ten-second FL wrist window and its recorded side.

Existing methods that draw on higher-quality references either use them at prediction time, as in sensor fusion \citep{ordonez2016}, or train a new wrist model by feature adaptation across placements \citep{chang2020,chen2019}, knowledge distillation \citep{hinton2015,lopezpaz2016} into a wrist-only student, or fine-tuning on labeled wrist data. Fusion needs the reference sensor at deployment. A new wrist model replaces the classifier trained on reference recordings and fits its activity boundaries to the blurred wrist signal. We propose WearableBridge, which keeps the reference encoder and classifier fixed and learns a small map, called the bridge, that moves wrist features within the reference feature space. The bridge needs only the wrist window at deployment, and the fixed classifier keeps the activity boundaries it learned from reference recordings, in which activities are well separated.

\begin{figure}[t]
\centering
\includegraphics[width=0.94\linewidth]{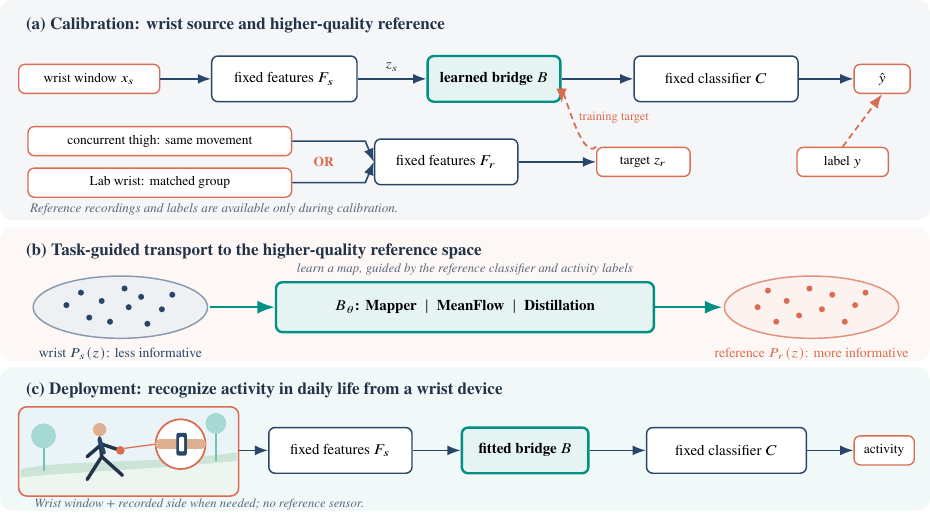}
\caption{WearableBridge overview. (a) Calibration pairs each wrist window with a higher-quality reference: a concurrent thigh window (Thigh2Wrist) or a Lab wrist window matched by participant, activity, wrist side, and device placement (Lab2Freeliving). (b) The bridge $B_\theta$ (Mapper, MeanFlow, or Distillation) moves wrist features, distributed as $P_s$, toward the reference feature distribution $P_r$ and is trained with a classification loss through the fixed classifier $C$. (c) Deployment uses only a free-living wrist window and its recorded side.}
\label{fig:pipeline}
\end{figure}

The key observation is that a classifier trained on reference recordings defines regions of feature space associated with each activity, and those regions remain meaningful when no reference accompanies a deployed window. WearableBridge therefore trains the bridge as a task-guided transport map that carries each wrist feature into the region of its activity. During calibration, reference features or predictions indicate where wrist features should move, and a classification loss through the frozen classifier penalizes wrist features that land in the region of another activity. At deployment, the fixed classifier reads the mapped FL wrist feature (Figure~\ref{fig:pipeline}). We compare three forms of reference guidance, instantiated as three bridges: Mapper matches reference features, Distillation matches the reference classifier's class probabilities, and MeanFlow learns a one-step flow from wrist toward reference features \citep{geng2025,geng2026}.

On the Physical Activity Assessment Using Wearable Sensors (PAAWS) dataset \citep{potter2025}, with 40 held-out FL participants and more than 1.5 million wrist windows per design, every bridge improves macro-F1 in every run of both designs. The strongest, MeanFlow, raises macro-F1 from 35.07\% to 55.98\% in Thigh2Wrist, closing 44\% of the gap to the thigh sensor itself, and from 54.00\% to 63.54\% in Lab2Freeliving. Class-level results show that each reference transfers the distinctions it observes best: thigh posture separates sitting from standing, and laboratory wrist sessions separate cycling from walking. On selfBACK, MeanFlow raises macro-F1 from 25.73\% to 66.74\%, above the thigh sensor itself (64.42\%). Fitted only on PAAWS, MeanFlow also raises macro-F1 on the CAPTURE-24 cohort from 42.75\% to 54.73\% without using its labels. The gains hold across four reference encoders and grow with temporal context, where MeanFlow reaches 61.70\% in Thigh2Wrist and 68.90\% in Lab2Freeliving.

Our contributions are:
\begin{itemize}[leftmargin=1.2em, itemsep=2pt, topsep=3pt, parsep=0pt]
    \item \textbf{A reference-reuse formulation} that casts wrist recognition with limited reference data as learning a map that moves wrist features into the activity regions of a fixed classifier trained on reference recordings. It covers instance-level (Thigh2Wrist) and group-level (Lab2Freeliving) references, and a sufficiently expressive map can reproduce the predictions of any classifier trained directly on wrist features (Section~\ref{sec:method}).
    \item \textbf{WearableBridge}, a learning objective that combines a classification loss through the frozen classifier with a reference-guidance term, instantiated as feature matching (Mapper), class-probability matching (Distillation), or a one-step flow conditioned on the wrist feature (MeanFlow).
    \item \textbf{Consistent gains for participants without reference recordings}: every bridge improves macro-F1 on PAAWS with all four reference encoders, and the gains replicate on selfBACK and transfer to CAPTURE-24 without target labels.
\end{itemize}

\section{Related Work}
\label{sec:related}
\noindent\textbf{Transfer across positions and protocols.} Domain adaptation aligns feature distributions across domains \citep{ganin2016,courty2017}, including wearable devices and placements \citep{chang2020,chen2019}, and ColloSSL learns representations from simultaneous unlabeled recordings across worn devices \citep{jain2022}. These methods train a new encoder or classifier for the target domain and replace the classifier trained on the higher-quality reference, together with the activity boundaries it learned. WearableBridge keeps that classifier and its encoder fixed and learns where wrist features should land among its activity regions.

\noindent\textbf{Privileged information and cross-modal distillation.} Knowledge distillation and learning with privileged information train a student from a teacher that sees more at training time \citep{hinton2015,lopezpaz2016}, including a teacher on a paired modality \citep{gupta2016}. The student is a new model, and each training input needs a paired teacher input. WearableBridge keeps the teacher's classifier fixed and trains only a small map of wrist features into its feature space. It also learns from group-matched references recorded in separate sessions, as in Lab2Freeliving, where no paired teacher input exists.

\noindent\textbf{Adapting frozen models.} Model reprogramming learns an input transformation that makes a frozen network perform a new task \citep{elsayed2019}, adapters train small layers inserted into a fixed network \citep{houlsby2019}, and test-time adaptation updates normalization parameters by minimizing prediction entropy on unlabeled test data \citep{wang2021tent}. Their losses score only the model's output. WearableBridge also transforms the input to a fixed classifier, here a wrist feature, and each wrist window comes with a reference recording of the same activity whose feature or prediction indicates where the wrist feature should land.

\noindent\textbf{Flow-based transport.} Flow matching and rectified flow learn velocity fields that carry one distribution to another \citep{lipman2023,liu2023rectified}, bridge models learn maps between paired domains \citep{liu2023i2sb}, and MeanFlow learns average velocities for one-step generation \citep{geng2025,geng2026}. These models are trained to reproduce a target distribution. Recognition requires each transported wrist feature to land in the decision region of its activity. WearableBridge conditions MeanFlow on the wrist feature and adds a classification loss through the frozen classifier, so the transport is guided by the task and takes one network evaluation at deployment.

\section{WearableBridge: Task-Guided Latent Transport for Free-Living Activity Recognition}
\label{sec:method}
The goal is to recognize activities from wrist accelerometer data recorded in free living. Labeled data of higher quality come from a reference source: a thigh sensor worn at the same time as the wrist sensor in Thigh2Wrist, and wrist recordings from a laboratory session in Lab2Freeliving. A classifier trained on reference features divides the reference feature space into one region per activity, but because wrist and reference recordings differ, a wrist feature can fall outside the region of its activity. WearableBridge keeps the reference encoder and classifier fixed and learns a bridge, a map within the reference feature space that carries each wrist feature into the region the classifier associates with its activity.

Section~\ref{sec:setup} defines the data, the fixed encoder and classifier, and the prediction rule, and shows that a bridge followed by the frozen classifier can reproduce the predictions of a classifier retrained on wrist features. Section~\ref{sec:objective} gives the training objective and three bridges, Mapper, Distillation, and MeanFlow, which differ in how they use the reference data.

\subsection{Problem setup}
\label{sec:setup} 
WearableBridge training uses a calibration study, in which each wrist window comes with a reference window and an activity label; prediction uses the wrist window alone.
We mark the wrist input, the source, by subscript $s$ and the higher-quality reference by subscript $r$. Let $x_s$ be a ten-second wrist accelerometer window, $y\in\mathcal{Y}$ its label in the activity set $\mathcal{Y}$, and $x_r$ an associated reference window. In Thigh2Wrist, $x_r$ is the thigh window recorded simultaneously with $x_s$. In Lab2Freeliving, $x_s$ is a SimFL wrist window and $x_r$ a Lab wrist window drawn from the same participant, activity, wrist side, and device placement. All wrist windows follow the left-wrist axis convention, so right-wrist windows have their first axis reversed (Appendix~\ref{app:pairing}).

One fixed encoder, written $F_s$ on wrist windows and $F_r$ on reference windows, maps both into the reference feature space $\mathbb{R}^d$. The reference classifier $C:\mathbb{R}^d\rightarrow\mathbb{R}^{|\mathcal{Y}|}$ is trained on labeled reference features and then frozen.

The bridge acts on standardized features. We standardize features with the coordinatewise mean $\mu$ and standard deviation $\sigma$ of training reference features (Appendix~\ref{app:pairing}). For a raw feature $h\in\mathbb{R}^d$, write $S(h)=(h-\mu)\oslash\sigma$ and its inverse $D(z)=\mu+\sigma\odot z$, with coordinatewise $\oslash$ and $\odot$, and define the standardized wrist and reference features $z_s=S(F_s(x_s))$ and $z_r=S(F_r(x_r))$. The bridge $B_\theta:\mathbb{R}^d\rightarrow\mathbb{R}^d$, with parameters $\theta$, maps $z_s$ within this space, and the classifier produces logits $a\in\mathbb{R}^{|\mathcal{Y}|}$ and a prediction $\hat y$ through
\begin{equation}
 a=C\!\left(D\!\left(B_\theta(z_s)\right)\right),
 \qquad \hat y=\arg\max_{k\in\mathcal{Y}}a_k.
\label{eq:prediction-rule}
\end{equation}
Only the bridge is trained, with gradients passing through the frozen classifier. Setting $B_\theta(z)=z$ gives the unadapted reference, our baseline. Every bridge is initialized at this identity map by zero-initializing its output layer, so training begins from the baseline predictions. Bridge fitting uses no data from held-out FL participants.

\paragraph{Why keep the classifier frozen?} The alternative to a bridge is to retrain a classifier on labeled wrist features. A sufficiently expressive bridge, followed by the frozen classifier, can reproduce any classifier trained directly on wrist features. Let $C_s:\mathbb{R}^d\rightarrow\mathbb{R}^{|\mathcal{Y}|}$ be such a classifier on standardized wrist features. Both calibration designs use a linear reference classifier $C(h)=Wh+c$. When $W\in\mathbb{R}^{|\mathcal{Y}|\times d}$ has full row rank, the bridge $B_\theta(z)=S\{W^{+}(C_s(z)-c)\}$, with $W^{+}$ the Moore--Penrose pseudoinverse, gives $C(D(B_\theta(z)))=WW^{+}(C_s(z)-c)+c=C_s(z)$ because $WW^{+}=I$, so it reproduces the logits of $C_s$ exactly. For a nonlinear reference classifier, a bridge reproduces the predicted labels of $C_s$ by sending each wrist feature into the region of the reference feature space that $C$ associates with the label $C_s$ assigns, which requires only that every activity has a nonempty region. Reusing $C$ therefore keeps every prediction available from retraining on wrist features, and reference guidance selects among these maps during training.

\subsection{Training objective and bridges}
\label{sec:objective}
The training objective uses the two kinds of supervision in the calibration study: the activity label, through a cross-entropy term shared by all three bridges, and the reference window, through a reference-guidance term that differs across them.
Let $\Pi$ denote the joint distribution of $(x_s,x_r,y)$ in the calibration study, where higher-quality references and activity labels are available. In Thigh2Wrist, $x_r$ is the thigh window recorded with $x_s$, and in Lab2Freeliving, it is drawn within the participant--activity--wrist-side group $g$ of $x_s$. All three bridges minimize a weighted empirical version of one unified objective,
\begin{equation}
 \mathcal{L}(\theta)=\mathbb{E}_{\Pi}\!\left[\rho_\theta(z_s,z_r)+\CE(a,y)\right],
 \qquad \CE(a,y)=-\log\softmax(a)_y,
\label{eq:unified}
\end{equation}
with $a$ from \eqref{eq:prediction-rule}. The cross-entropy term $\CE$ passes gradients through the frozen classifier and moves each wrist feature toward the decision region $\{z:\arg\max_k C(D(z))_k=y\}$ of its activity. The classifier learned these regions from reference features, and the reference-guidance term $\rho_\theta$ indicates where within them a wrist feature should land. The three bridges specify $\rho_\theta$ through reference features (Mapper), reference class probabilities (Distillation), and a path between paired features (MeanFlow). 

\paragraph{Mapper.} The reference feature supplies an endpoint target:
\begin{equation}
 \ell_{\mathrm{map}}=\frac{1}{d}\|B_\theta(z_s)-z_r\|_2^2+\CE(a,y).
\label{eq:mapper}
\end{equation}
In Thigh2Wrist, $z_r$ is the concurrent thigh feature. In Lab2Freeliving, the Lab window paired with each SimFL window is redrawn within its group $g$ at every epoch, so $z_r$ is approximately independent of $z_s$ given $g$. Because $B_\theta(z_s)$ depends only on $z_s$, the expected squared error then equals $\mathbb{E}\|B_\theta(z_s)-m(z_s)\|_2^2$ plus a term that does not depend on $\theta$, with $m(z_s)=\mathbb{E}[\mathbb{E}(z_r\mid g)\mid z_s]$. Mapper therefore moves each wrist feature toward reference group means, which act as activity prototypes.

\paragraph{Distillation.} The reference classifier supplies a probability target. Let $q_r=\softmax(C(D(z_r))/T)$ and $q_s=\softmax(a/T)$ be the class probabilities of the reference feature and the mapped wrist feature at temperature $T$, with $T=2$ in all experiments. The loss is
\begin{equation}
 \ell_{\mathrm{dist}}=T^2\KL(q_r\,\|\,q_s)+\CE(a,y),
\label{eq:distill}
\end{equation}
where $\KL$ denotes Kullback--Leibler divergence. The bridge matches the reference classifier's predictions, so the mapped wrist feature can differ from the reference feature whenever their class probabilities agree. In Lab2Freeliving, the expected loss fits $q_s$ to the group-averaged soft label $\mathbb{E}[q_r\mid z_s]$, up to a term that does not depend on $\theta$, so Distillation targets the group's average class probabilities.

\paragraph{Source-conditioned MeanFlow.} The path between paired features supplies a velocity target, and the bridge moves a wrist feature from the wrist end of this path to the reference end in one step. We adapt MeanFlow~\citep{geng2025,geng2026} to learn this one-step update conditioned on the wrist feature. For each calibration pair, define a straight path from the reference feature at time zero to the wrist feature at time one, with constant velocity $v$:
\begin{equation}
 z(\tau)=(1-\tau)z_r+\tau z_s,\qquad v=z_s-z_r.
\label{eq:path}
\end{equation}
For $0\leq t_0\leq t\leq1$, the network $u_\theta(z(t),t_0,t\mid z_s)$ predicts the average velocity over the interval $[t_0,t]$, so that an exact prediction gives $z(t_0)=z(t)-(t-t_0)\,u_\theta$. Conditioning on $z_s$ gives the network the original wrist feature at every point on the path. At deployment, one network evaluation carries the wrist feature from time one back to time zero:
\begin{equation}
 B_\theta(z_s)=z_s-u_\theta(z_s,0,1\mid z_s).
\label{eq:one-step}
\end{equation}
Differentiating $(t-t_0)\,u_\theta=z(t)-z(t_0)$ in $t$ shows that an exact prediction satisfies $u_\theta+(t-t_0)\,du_\theta/dt=\dot z(t)=v$, and training fits $\theta$ to this identity. Following \citet{geng2026}, $du_\theta/dt$ is computed by the chain rule with $t_0$ and $z_s$ held fixed and with the predicted instantaneous velocity $u_{\mathrm{inst}}=u_\theta(z(t),t,t\mid z_s)$ in place of $\dot z(t)$:
\begin{align}
 V_\theta&=u_\theta+(t-t_0)\,\sg\!\left[\partial_tu_\theta+
 (\nabla_zu_\theta)\,u_{\mathrm{inst}}\right],\label{eq:mf-residual}\\
 e&=\|V_\theta-v\|_2^2,\qquad
 \ell_{\mathrm{path}}=\frac{e}{\sg(e)+\epsilon},\label{eq:mf-loss}
\end{align}
where $u_\theta=u_\theta(z(t),t_0,t\mid z_s)$, the stop-gradient $\sg$ treats its argument as a constant, and $\epsilon>0$ stabilizes the adaptive weight. Half of each batch uses $t_0=t$, so that $V_\theta=u_{\mathrm{inst}}$ is regressed directly onto $v$ (Appendix~\ref{app:meanflow-sampling}).

The complete objective adds source preservation and the cross-entropy term at the one-step output:
\begin{equation}
 \ell_{\mathrm{flow}}=\ell_{\mathrm{path}}+
 \frac{0.01}{d}\|B_\theta(z_s)-z_s\|_2^2+\CE(a,y).
\label{eq:full-mf}
\end{equation}
The source-preservation term keeps updates small, and the cross-entropy term trains the one-step update for the fixed classifier.

As a summary, in the notation of \eqref{eq:unified}, $\rho_\theta$ is the squared error scaled by $1/d$ for Mapper, the KL divergence scaled by $T^2$ for Distillation, and, for MeanFlow, the path loss $\ell_{\mathrm{path}}$ averaged over sampled times plus source preservation. Each bridge minimizes a batch mean of weighted per-pair losses (Appendix~\ref{app:architecture}).

\providecommand{\sd}[1]{{\scriptsize$\pm$#1}}

\section{Empirical Studies}
\label{sec:experiments}

We evaluate WearableBridge on the PAAWS dataset in both calibration designs, Thigh2Wrist and Lab2Freeliving, and on two external datasets, selfBACK and CAPTURE-24. The studies ask four questions. Does bridging improve wrist-only activity recognition? Does the improvement depend on the encoder, the activity set, or temporal context? Does it carry to other cohorts? How does reusing the frozen reference classifier compare with retraining a classifier on wrist features?

We find that bridging improves recognition in both designs (Section~\ref{sec:exp-main}). MeanFlow, the best bridge, raises macro-F1 from 35.07\% to 55.98\% in Thigh2Wrist, closing 44\% of the gap to the measured thigh, and from 54.00\% to 63.54\% in Lab2Freeliving. Each reference transfers the distinctions it observes best: the thigh separates the postures that the wrist confuses, and the Lab wrist separates cycling from walking. The gains hold for three reference encoders trained from scratch and with six activities, and averaging over neighboring windows adds a further 5.26 to 6.15 percentage points to every bridge (Section~\ref{sec:exp-robust}). They replicate on selfBACK, where MeanFlow raises macro-F1 from 25.73\% to 66.74\%, and a bridge fitted on PAAWS raises macro-F1 on CAPTURE-24 from 42.75\% to 54.73\% without target labels or refitting (Section~\ref{sec:exp-external}). Reusing the reference classifier is competitive with retraining: MeanFlow exceeds the Direct wrist classifier on PAAWS Thigh2Wrist and selfBACK and trails it by 0.85 points on CAPTURE-24. Reference guidance adds 2.77 points over the activity-only bridge on PAAWS Thigh2Wrist, where the thigh is far more informative than the wrist, but not on selfBACK, where the two sensors are about equally informative (Section~\ref{sec:exp-retrain}).
\raggedbottom
\subsection{Experimental setup}
\label{sec:exp-setup}

\noindent\textbf{Data.} We use the PAAWS dataset \citep{potter2025}. Calibration draws on Lab and SimFL recordings from 170 training and 43 validation participants, and testing uses FL recordings from 40 held-out participants. Thigh2Wrist uses concurrent wrist and thigh windows from both sessions. The primary PAAWS comparisons classify cycling, sitting, standing, and walking from ten-second wrist windows at 50\,Hz. All four activities have labeled examples in Lab, SimFL, and FL.

Experiments are also run on two external datasets. selfBACK \citep{selfback2020dataset} provides a second Thigh2Wrist cohort, with concurrent wrist and thigh recordings of nine activities, including sitting, standing, and walking, from 20 training, six validation, and seven test participants. CAPTURE-24 \citep{chan2024capture24} tests transfer to a new cohort with 134,658 free-living wrist windows of the four PAAWS activities from 50 participants. All fitting and selection use PAAWS, so CAPTURE-24 serves only for evaluation (Appendices~\ref{app:pairing}, \ref{app:selfback-setup}, and~\ref{app:capture-sourceonly}).

\begin{table}[!htbp]
\setlength{\belowcaptionskip}{6pt}
\centering
\caption{PAAWS current-window results with frozen MoCA references, in percent (mean $\pm$ sample SD over three runs). The unadapted reference is the wrist-input baseline: it applies the frozen reference classifier directly to wrist features. Mapper, MeanFlow, and Distillation are the WearableBridge variants, which map wrist features before the same classifier. Measured thigh, below the last rule, applies the same classifier to the concurrent thigh window at test time and shows the accuracy of the reference itself. Both designs test the same 40 participants, and bold marks the highest wrist-input mean per column.}
\label{tab:main}
\label{tab:wrist-thigh}
\footnotesize
\setlength{\tabcolsep}{1.2pt}
{\raggedright\small\textbf{(a) Thigh2Wrist.} 1,588,369 test windows.\par}
\smallskip
\begin{tabular*}{\linewidth}{@{\extracolsep{\fill}}lccccccc@{}}
\toprule
 & \multicolumn{3}{c}{Overall} & \multicolumn{4}{c}{Class F1}\\
\cmidrule(lr){2-4}\cmidrule(l){5-8}
Method & Macro-F1 & Bal.\ acc. & Accuracy & Cycling & Sitting & Standing & Walking\\
\midrule
Unadapted reference {\scriptsize(baseline)} & 35.07\sd{0.21} & 42.45\sd{0.13} & 51.80\sd{0.30} & 2.77\sd{0.03} & 70.27\sd{0.53} & 18.13\sd{0.24} & 49.12\sd{1.11}\\
\midrule
Mapper & 54.13\sd{0.54} & 67.00\sd{1.91} & 73.83\sd{1.93} & 17.14\sd{3.81} & 83.82\sd{1.72} & 35.69\sd{0.48} & 79.87\sd{0.40}\\
MeanFlow & \textbf{55.98}\sd{0.51} & \textbf{69.87}\sd{0.73} & \textbf{76.11}\sd{1.64} & \textbf{19.01}\sd{2.56} & \textbf{85.66}\sd{1.41} & \textbf{38.07}\sd{0.50} & \textbf{81.17}\sd{0.11}\\
Distillation & 55.06\sd{0.32} & 66.35\sd{0.89} & 75.93\sd{1.85} & 17.84\sd{2.43} & 85.52\sd{1.55} & 36.78\sd{0.16} & 80.12\sd{0.48}\\
\midrule
Measured thigh {\scriptsize(reference sensor)} & 82.50\sd{0.33} & 89.62\sd{0.10} & 94.97\sd{0.01} & 61.81\sd{1.45} & 97.97\sd{0.02} & 82.87\sd{0.12} & 87.37\sd{0.08}\\
\bottomrule
\end{tabular*}\par
\medskip
{\raggedright\small\textbf{(b) Lab2Freeliving.} 1,595,771 test windows. \par}
\smallskip
\begin{tabular*}{\linewidth}{@{\extracolsep{\fill}}lccccccc@{}}
\toprule
 & \multicolumn{3}{c}{Overall} & \multicolumn{4}{c}{Class F1}\\
\cmidrule(lr){2-4}\cmidrule(l){5-8}
Method & Macro-F1 & Bal.\ acc. & Accuracy & Cycling & Sitting & Standing & Walking\\
\midrule
Unadapted reference {\scriptsize(baseline)} & 54.00\sd{1.29} & 62.34\sd{1.93} & 78.00\sd{2.18} & 9.42\sd{2.76} & 87.49\sd{1.64} & \textbf{41.25}\sd{1.15} & 77.85\sd{1.37}\\
\midrule
Mapper & 60.90\sd{0.57} & 65.88\sd{1.15} & 75.63\sd{2.01} & 43.60\sd{0.72} & 85.22\sd{1.61} & 34.48\sd{1.10} & 80.28\sd{0.39}\\
MeanFlow & \textbf{63.54}\sd{1.23} & \textbf{67.93}\sd{0.94} & \textbf{80.02}\sd{1.47} & \textbf{47.21}\sd{2.21} & \textbf{88.58}\sd{1.06} & 38.09\sd{1.16} & \textbf{80.29}\sd{0.79}\\
Distillation & 62.30\sd{1.26} & 66.80\sd{0.89} & 79.05\sd{1.55} & 44.77\sd{2.86} & 87.91\sd{1.13} & 36.30\sd{1.00} & 80.20\sd{0.52}\\
\bottomrule
\end{tabular*}
\end{table}

\noindent\textbf{Models.} The primary encoder is the Multi-modal Cross-masked Autoencoder (MoCA) \citep{ryu2025}, a masked-autoencoder model for multichannel wearable time series with public pretrained weights. Its features are learned without activity labels, and one frozen network embeds wrist, thigh, and Lab windows in the shared space that the formulation requires. It supplies features for a linear thigh classifier in Thigh2Wrist and for a Lab-trained projection and classifier in Lab2Freeliving (Appendix~\ref{app:architecture}). To test whether gains depend on MoCA, we repeat both designs with DeepConvLSTM, InceptionTime, and HART reference encoders trained from scratch on PAAWS training participants \citep{ordonez2016,ismailfawaz2020,ek2023}. CAPTURE-24 uses HARNet10, pretrained on free-living wrist accelerometry from UK Biobank \citep{yuan2024harnet}.

\noindent\textbf{Comparators.} Additional comparisons train on the same labeled calibration wrist windows, which fixes the data budget. The Direct wrist classifier tests reuse against retraining: it freezes the encoder and retrains the reference classifier on wrist features. The activity-only bridge tests reference guidance: it keeps the MeanFlow architecture and trains with the cross-entropy term alone, dropping both the path loss and source preservation ($\rho_\theta\equiv0$ in \eqref{eq:unified}). Joint encoder fine-tuning is reported separately in Appendix~\ref{app:joint-finetuning}.

\noindent\textbf{Evaluation.} Macro-F1 is the primary metric because it weights activities equally to address the imbalance of classes in the data. Balanced accuracy, accuracy, and class F1 are secondary, and differences are in percentage points (pp), computed before rounding. We report the mean and sample SD over three runs with different random seeds and a fixed participant split. Paired 95\% participant-bootstrap intervals for all differences are reported in Appendix~\ref{app:evaluation} (Table~\ref{tab:bootstrap}) and Appendix~\ref{app:six-class} (Table~\ref{tab:six-class}). Predictions use only the current window without additional context given by neighboring windows, unless stated otherwise.

\begin{figure}[!h]
\centering
\includegraphics[width=\linewidth]{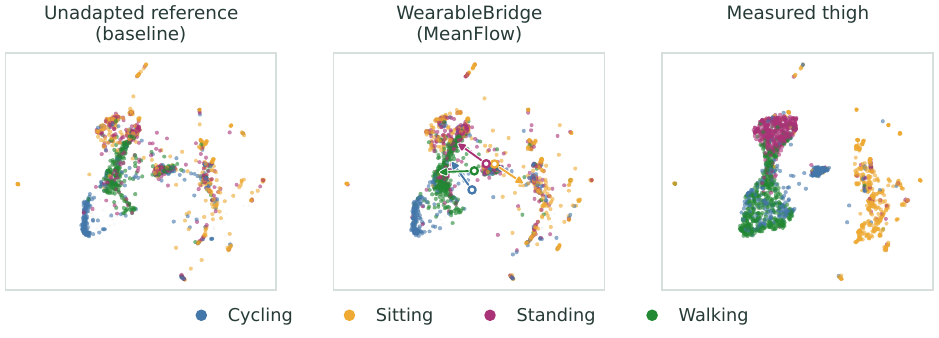}
\caption{Thigh2Wrist features before and after MeanFlow, with concurrent measured thighs. A fixed UMAP \citep{mcinnes2018} fitted on SimFL-validation thigh features projects 2,048 held-out FL wrist--thigh pairs, sampled equally by activity and wrist side, with validation references in faint gray. Class means are computed over 8,192 pairs. Open circles mark each activity's mean wrist input, and arrows point toward its mean output, with lengths scaled for readability (Appendix~\ref{app:visualization}).}
\label{fig:thigh2wrist-umap}
\end{figure}

\begin{figure}[!h]
\centering
\includegraphics[width=\linewidth]{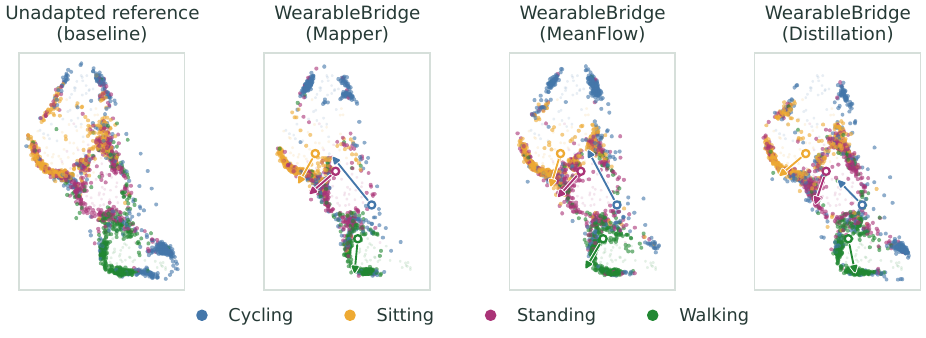}
\caption{Lab2Freeliving features before and after each bridge. A fixed UMAP \citep{mcinnes2018} fitted on Lab-validation references (faint points) projects 2,048 held-out FL wrist windows, sampled equally by activity and wrist side, and their mapped outputs. Circles and arrows are as in Figure~\ref{fig:thigh2wrist-umap}. \interactivefigures}
\label{fig:umap}
\label{fig:lab2freeliving-umap}
\end{figure}

\subsection{Bridging improves wrist-only recognition in both designs}
\label{sec:exp-main}
Every bridge improves macro-F1 over the baseline in every run of both calibration designs (Table~\ref{tab:main}). MeanFlow has the highest mean macro-F1, balanced accuracy, and accuracy in each design.

In Thigh2Wrist, MeanFlow raises macro-F1 from 35.07\% to 55.98\%, a gain of 20.91 percentage points. This closes 44\% of the gap to the measured thigh (82.50\%), which applies the same classifier to the concurrent thigh window. The gain extends to all four activities: walking F1 rises from 49.12\% to 81.17\%, sitting F1 from 70.27\% to 85.66\%, standing F1 from 18.13\% to 38.07\%, and cycling F1, the rarest class, nearly sevenfold, from 2.77\% to 19.01\%. In Lab2Freeliving, MeanFlow raises macro-F1 from 54.00\% to 63.54\%, a gain of 9.54 percentage points from group-matched Lab references. Most of the gain comes from cycling, whose F1 rises from 9.42\% to 47.21\%, walking F1 rises from 77.85\% to 80.29\%, and standing F1 dips from 41.25\% to 38.09\%.

An interesting heuristic is revealed: \textbf{each reference transfers the distinctions it observes best}. The thigh records leg posture and motion, which separate the four activities far more clearly than arm movement does (Table~\ref{tab:main}). In Figure~\ref{fig:thigh2wrist-umap} (Thigh2Wrist), MeanFlow moves the mean wrist features of sitting, standing, and walking toward the corresponding thigh regions. The largest correction removes false cycling predictions: the unadapted reference labels 56.6\% of walking windows and 38.7\% of standing windows as cycling, and MeanFlow lowers these shares to 1.4\% and 4.1\% (Appendix~\ref{app:additional}). Sitting and standing, which the wrist confuses because arm movements are similar in both postures, lie far apart in the thigh space, and their recalls rise from 56.3\% to 78.7\% and from 32.9\% to 56.0\%. Standing and walking are both upright and differ mainly in thigh motion, which also varies while standing, so they lie close together and account for most remaining walking errors. 

In Figure~\ref{fig:lab2freeliving-umap} (Lab2Freeliving), all three bridges move each activity's mean wrist feature in similar directions, consistent with the group-level targets in Section~\ref{sec:method}. Cycling windows, which the unadapted reference places near walking, move toward the Lab cycling region, and the share of cycling windows predicted as walking falls from 46.3\% to 12.0\% with MeanFlow (Appendix~\ref{app:additional}). This pattern is consistent with Lab wrist sessions recording arm posture and motion cleanly, which separates cycling, with the hands on the handlebars, from walking with its arm swing. Sitting and standing look alike at the wrist, so the Lab reference offers little to separate them: standing windows predicted as sitting rise from 34.1\% to 39.1\%, which lowers standing recall from 54.4\% to 51.1\%.

\subsection{Gains extend across encoders, activity classes, and temporal context}
\label{sec:exp-robust}
\paragraph{Reference encoders.} The gains hold with DeepConvLSTM, InceptionTime, and HART reference encoders trained from scratch (Appendix Table~\ref{tab:backbones}). Every bridge improves mean macro-F1 over its unadapted-reference baseline by 9.62 to 20.15 percentage points in Thigh2Wrist, with improvements in every run, and by 8.13 to 12.76 in Lab2Freeliving. In Lab2Freeliving, the three bridges are within 0.7 percentage points of each other for every encoder. These encoders and their classifiers learn only from reference windows of PAAWS training participants, so WearableBridge also improves reference models built entirely within the calibration study.

\paragraph{Six-class Thigh2Wrist.} With lying and stairs added, MeanFlow raises macro-F1 from 14.46\% for the unadapted reference to 44.04\% and exceeds the Direct wrist classifier (40.55\%) by 3.49 percentage points. It also improves on the Direct wrist classifier in F1 for all six activities and in macro-F1 for 39 of 40 participants (Appendix~\ref{app:six-class}). Lab2Freeliving pairs SimFL with Lab windows, and among the calibration sessions only Lab includes lying, so the extension uses Thigh2Wrist.

\paragraph{Temporal context.}
Averaging class probabilities over the current window and up to four preceding windows (five-window averaging) improves every bridge in both designs by 5.26 to 6.15 percentage points in macro-F1, and MeanFlow reaches 61.70\% in Thigh2Wrist and 68.90\% in Lab2Freeliving (Appendix Table~\ref{tab:temporal}). Without a bridge, averaging adds only 0.46 points in Thigh2Wrist, because the unadapted reference errs consistently, as in its false cycling predictions (Section~\ref{sec:exp-main}), and neighboring windows share the error. The bridge removes this shift, and averaging then corrects many of the remaining isolated errors, so bridging and temporal context are complementary.

\subsection{Gains replicate on selfBACK and transfer to CAPTURE-24}
\label{sec:exp-external}
\label{sec:selfback-paired}
\label{sec:capture-sourceonly}
On selfBACK, the reference is a nine-class thigh classifier on frozen MoCA features, and the seven test participants follow structured sessions (Table~\ref{tab:external}). MeanFlow raises macro-F1 from 25.73\% to 66.74\%, a gain of 41.01 percentage points, and all three bridges exceed 65\%. On this dataset the wrist is nearly as informative as the thigh (Direct wrist classifier 63.59\%, measured thigh 64.42\%), and every bridge, fitted with activity labels through the thigh classifier, exceeds both.

\begin{table}[!htbp]
\setlength{\belowcaptionskip}{6pt}
\centering
\caption{selfBACK Thigh2Wrist replication, in percent (mean $\pm$ sample SD over three runs), on 963 wrist--thigh test windows from seven participants. The unadapted reference is the wrist-input baseline, and comparators are defined in Section~\ref{sec:exp-setup}. Measured thigh applies the thigh classifier to the concurrent thigh window at test time.}
\label{tab:external}
\label{tab:selfback-paired}
\footnotesize
\setlength{\tabcolsep}{1.2pt}
\begin{tabular*}{\linewidth}{@{\extracolsep{\fill}}lccccccc@{}}
\toprule
 & Baseline & \multicolumn{3}{c}{WearableBridge} & \multicolumn{2}{c}{Comparators} & Reference sensor\\
\cmidrule(lr){2-2}\cmidrule(lr){3-5}\cmidrule(lr){6-7}\cmidrule(l){8-8}
Metric & Unadapted & Mapper & MeanFlow & Distillation & Activity-only & Direct wrist & Measured thigh\\
\midrule
Macro-F1 & 25.73\sd{0.19} & 65.80\sd{1.04} & 66.74\sd{0.38} & 65.43\sd{1.06} & 67.76\sd{1.34} & 63.59\sd{0.07} & 64.42\sd{0.37}\\
Bal.\ acc. & 32.96\sd{0.09} & 67.44\sd{0.41} & 68.11\sd{0.19} & 67.18\sd{0.71} & 68.93\sd{1.58} & 64.62\sd{0.14} & 65.51\sd{0.21}\\
Accuracy & 35.31\sd{0.10} & 70.61\sd{1.28} & 71.37\sd{0.42} & 70.54\sd{1.17} & 71.89\sd{1.53} & 68.78\sd{0.60} & 67.88\sd{0.22}\\
\bottomrule
\end{tabular*}
\end{table}

MeanFlow is fitted on PAAWS in the Lab2Freeliving design, with the HARNet10 encoder and PAAWS Lab classifier fixed, and applied to CAPTURE-24 without target labels or refitting. It raises macro-F1 from 42.75\% to 54.73\%, a gain of 11.98 percentage points, and cycling F1 rises from 5.38\% to 50.64\% (Appendix Table~\ref{tab:capture-sourceonly}). As in PAAWS Lab2Freeliving, the gain concentrates on cycling, so the correction learned from PAAWS SimFL wrist data carries to free-living recordings in a new cohort.

\subsection{Reference reuse is competitive with the Direct wrist classifier}
\label{sec:exp-retrain}
\textbf{Can the frozen reference classifier serve wrist features as well as a classifier retrained on them? Yes.} In macro-F1, MeanFlow matches or exceeds the Direct wrist classifier trained on the same wrist labels. Reuse also keeps one reference model for all participants, so predictions for wrist-only participants and for participants with reference recordings share the same activity boundaries. Reference recordings further guide training, which adds accuracy where the reference is far more informative than the wrist.

\noindent\textbf{Direct wrist classifier.} MeanFlow exceeds the Direct wrist classifier on PAAWS Thigh2Wrist, 55.98\% versus 54.60\% (Table~\ref{tab:wrist-controls}), and on selfBACK by 3.16 percentage points (Table~\ref{tab:external}). On CAPTURE-24 (Appendix Table~\ref{tab:capture-sourceonly}), where both are fitted on labeled PAAWS SimFL windows, MeanFlow is within 0.85 percentage points of the Direct wrist classifier in macro-F1 (54.73\% versus 55.58\%), and the Direct wrist classifier has higher balanced accuracy and accuracy by 7.67 and 2.77 percentage points.

\noindent\textbf{Activity-only bridge.} On PAAWS Thigh2Wrist, where the measured thigh reaches 82.50\%, MeanFlow exceeds the activity-only bridge by 2.77 percentage points (Table~\ref{tab:wrist-controls}). The two bridges share architecture, labels, the thigh classifier, and feature standardization, so this gain measures what the reference-guidance term $\rho_\theta$ adds. On selfBACK, where the measured thigh (64.42\%) scores below every bridge, the activity-only bridge is highest (67.76\%), so the bridge itself carries the gain.

\begin{table}[!htbp]
\setlength{\belowcaptionskip}{6pt}
\centering
\caption{PAAWS Thigh2Wrist comparisons with a frozen encoder on the same FL test windows as Table~\ref{tab:main}. Values are percentages (mean $\pm$ sample SD over three runs). All methods train on the same labeled wrist windows, and comparators are defined in Section~\ref{sec:exp-setup}. Bold marks the highest mean per column.}
\label{tab:wrist-controls}
\footnotesize
\setlength{\tabcolsep}{1.2pt}
\begin{tabular*}{\linewidth}{@{\extracolsep{\fill}}lccccccc@{}}
\toprule
 & \multicolumn{3}{c}{Overall} & \multicolumn{4}{c}{Class F1}\\
\cmidrule(lr){2-4}\cmidrule(l){5-8}
Method & Macro-F1 & Bal.\ acc. & Accuracy & Cycling & Sitting & Standing & Walking\\
\midrule
Direct wrist classifier & 54.60\sd{0.29} & 69.61\sd{0.15} & \textbf{76.56}\sd{0.47} & 15.11\sd{1.07} & \textbf{86.31}\sd{0.38} & 37.94\sd{0.47} & 79.04\sd{0.24}\\
Activity-only bridge & 53.21\sd{1.08} & 64.81\sd{0.65} & 73.29\sd{1.99} & 17.07\sd{2.36} & 83.46\sd{1.63} & 34.12\sd{1.25} & 78.20\sd{0.27}\\
MeanFlow & \textbf{55.98}\sd{0.51} & \textbf{69.87}\sd{0.73} & 76.11\sd{1.64} & \textbf{19.01}\sd{2.56} & 85.66\sd{1.41} & \textbf{38.07}\sd{0.50} & \textbf{81.17}\sd{0.11}\\
\bottomrule
\end{tabular*}
\end{table}

\section{Discussion and Conclusion}
\paragraph{Summary.} We asked how limited reference data can improve wrist-only recognition for participants without reference recordings. WearableBridge answers by reusing the reference model: a small bridge carries wrist features into the activity regions of the fixed reference classifier. Every bridge improves macro-F1 in every run of both PAAWS designs, and the gains hold across four encoders, replicate on selfBACK, and transfer to CAPTURE-24 without target labels. Because the reference model stays fixed, wrist-only participants and participants with reference recordings share one set of activity boundaries.

\paragraph{What each reference contributes.} Each reference transfers the distinctions it observes best. The thigh records leg posture and motion, which removes false cycling predictions and separates sitting from standing, and Lab wrist sessions separate cycling from walking. Reference guidance improves on the activity-only bridge on PAAWS Thigh2Wrist, where the thigh classifier is far more accurate than the bridged wrist, and the bridge itself carries the gains on selfBACK, where the two are comparable. In the primary four-class MoCA comparisons, the three bridges score within 2.7 percentage points of each other, consistent with their unified objective, and MeanFlow has the highest mean macro-F1 in both PAAWS designs.

\paragraph{Limitations and future work.} PAAWS calibration draws on 213 participants, and selfBACK calibration with 20 training participants already gains 41.01 percentage points. Subsampling PAAWS participants would show how gains scale with the amount of reference data. Per-term ablations with repeated runs would separate the effects of bridge architecture and loss. Combining thigh and Lab references, which separate different activity pairs, could improve all four activities, and reference guidance could extend to encoder fine-tuning and to several devices sharing one classifier.

\paragraph{Conclusion.} A calibration subset with reference recordings improves wrist-only recognition for every other participant through one fixed reference model. Existing Lab sessions serve as references without new data collection, and a bridge fitted in one study carries to another cohort.

\section*{Acknowledgments}

JZ's work was supported by NIH/NHLBI R01HL166802. YW's work was supported in part by funding from the Office of Naval Research under grant N00014-23-1-2590, the Department of Energy under grant DE-SC0026902, the National Science Foundation under grant DMS-2310831, DMS-2428059, IIS-2435696, IIS-2440954, Two Sigma Investments LP, Science of Trustworthy AI program at Schmidt Sciences, and LG Management Development Institute AI Research.

\clearpage
\bibliographystyle{abbrvnat}
\bibliography{references}

@article{chang2020,
  author = {Youngjae Chang and Akhil Mathur and Anton Isopoussu and Junehwa Song and Fahim Kawsar},
  title = {A Systematic Study of Unsupervised Domain Adaptation for Robust Human-Activity Recognition},
  journal = {Proceedings of the ACM on Interactive, Mobile, Wearable and Ubiquitous Technologies},
  volume = {4}, number = {1}, pages = {39:1--39:30}, year = {2020},
  doi = {10.1145/3380985}
}

@article{chen2019,
  author = {Yiqiang Chen and Jindong Wang and Meiyu Huang and Han Yu},
  title = {Cross-position activity recognition with stratified transfer learning},
  journal = {Pervasive and Mobile Computing}, volume = {57}, pages = {1--13}, year = {2019},
  doi = {10.1016/j.pmcj.2019.04.004}
}

@article{doherty2017,
  author = {Aiden Doherty and Dan Jackson and Nils Hammerla and Thomas Pl{\"o}tz and Patrick Olivier and Malcolm H. Granat and Tom White and Vincent T. van Hees and Michael I. Trenell and Christopher G. Owen and Stephen J. Preece and Rob Gillions and Simon Sheard and Tim Peakman and Soren Brage and Nicholas J. Wareham},
  title = {Large scale population assessment of physical activity using wrist worn accelerometers: The {UK Biobank} study},
  journal = {PLOS ONE}, volume = {12}, number = {2}, pages = {e0169649}, year = {2017},
  doi = {10.1371/journal.pone.0169649}
}

@article{ek2023,
  author = {Sannara Ek and Fran{\c c}ois Portet and Philippe Lalanda},
  title = {Transformer-based models to deal with heterogeneous environments in human activity recognition},
  journal = {Personal and Ubiquitous Computing}, volume = {27}, number = {6}, pages = {2267--2280}, year = {2023},
  doi = {10.1007/s00779-023-01776-3}
}

@inproceedings{geng2025,
  author = {Zhengyang Geng and Mingyang Deng and Xingjian Bai and J. Zico Kolter and Kaiming He},
  title = {Mean Flows for One-step Generative Modeling},
  booktitle = {Advances in Neural Information Processing Systems}, volume = {38}, year = {2025},
  pages = {75460--75482},
  doi = {10.52202/085713-2534}
}

@inproceedings{geng2026,
  author = {Zhengyang Geng and Yiyang Lu and Zongze Wu and Eli Shechtman and J. Zico Kolter and Kaiming He},
  title = {Improved Mean Flows: On the Challenges of Fastforward Generative Models},
  booktitle = {Proceedings of the IEEE/CVF Conference on Computer Vision and Pattern Recognition},
  pages = {30467--30476}, year = {2026}
}

@article{gyllensten2011,
  author = {Illapha Cuba Gyllensten and Alberto G. Bonomi},
  title = {Identifying Types of Physical Activity With a Single Accelerometer: Evaluating Laboratory-Trained Algorithms in Daily Life},
  journal = {IEEE Transactions on Biomedical Engineering}, volume = {58}, number = {9}, pages = {2656--2663}, year = {2011},
  doi = {10.1109/TBME.2011.2160723}
}

@misc{hinton2015,
  author = {Geoffrey Hinton and Oriol Vinyals and Jeffrey Dean},
  title = {Distilling the Knowledge in a Neural Network}, year = {2015},
  howpublished = {arXiv:1503.02531}, doi = {10.48550/arXiv.1503.02531}
}

@article{ismailfawaz2020,
  author = {Ismail Fawaz, Hassan and Benjamin Lucas and Germain Forestier and Charlotte Pelletier and Daniel F. Schmidt and Jonathan Weber and Geoffrey I. Webb and Lhassane Idoumghar and Pierre-Alain Muller and Fran{\c c}ois Petitjean},
  title = {{InceptionTime}: Finding {AlexNet} for Time Series Classification},
  journal = {Data Mining and Knowledge Discovery}, volume = {34}, number = {6}, pages = {1936--1962}, year = {2020},
  doi = {10.1007/s10618-020-00710-y}
}

@article{jain2022,
  author = {Yash Jain and Chi Ian Tang and Chulhong Min and Fahim Kawsar and Akhil Mathur},
  title = {{ColloSSL}: Collaborative Self-Supervised Learning for Human Activity Recognition},
  journal = {Proceedings of the ACM on Interactive, Mobile, Wearable and Ubiquitous Technologies}, volume = {6}, number = {1}, pages = {17:1--17:28}, year = {2022},
  doi = {10.1145/3517246}
}

@inproceedings{lopezpaz2016,
  author = {David Lopez-Paz and L{\'e}on Bottou and Bernhard Sch{\"o}lkopf and Vladimir Vapnik},
  title = {Unifying Distillation and Privileged Information},
  booktitle = {International Conference on Learning Representations}, year = {2016},
  url = {https://arxiv.org/abs/1511.03643}
}

@misc{mcinnes2018,
  author = {Leland McInnes and John Healy and James Melville},
  title = {{UMAP}: Uniform Manifold Approximation and Projection for Dimension Reduction},
  howpublished = {arXiv:1802.03426}, year = {2018}, doi = {10.48550/arXiv.1802.03426}
}

@article{ordonez2016,
  author = {Francisco Javier Ord{\'o}{\~n}ez and Daniel Roggen},
  title = {Deep Convolutional and {LSTM} Recurrent Neural Networks for Multimodal Wearable Activity Recognition},
  journal = {Sensors}, volume = {16}, number = {1}, pages = {115}, year = {2016},
  doi = {10.3390/s16010115}
}

@article{pavey2017,
  author = {Toby G. Pavey and Nicholas D. Gilson and Sjaan R. Gomersall and Bronwyn Clark and Stewart G. Trost},
  title = {Field evaluation of a random forest activity classifier for wrist-worn accelerometer data},
  journal = {Journal of Science and Medicine in Sport}, volume = {20}, number = {1}, pages = {75--80}, year = {2017},
  doi = {10.1016/j.jsams.2016.06.003}
}

@article{potter2025,
  author = {Veronika Potter and Hoan Tran and Daniel Mobley and Suzanne M. Bertisch and Dinesh John and Stephen Intille},
  title = {The Physical Activity Assessment Using Wearable Sensors ({PAAWS}) Dataset: Labeled Laboratory and Free-Living Accelerometer Data},
  journal = {Proceedings of the ACM on Interactive, Mobile, Wearable and Ubiquitous Technologies}, volume = {9}, number = {4}, pages = {204:1--204:32}, year = {2025},
  doi = {10.1145/3770639}
}

@misc{ryu2025,
  author = {Howon Ryu and Yuliang Chen and Yacun Wang and Andrea Z. LaCroix and Chongzhi Di and Loki Natarajan and Yu Wang and Jingjing Zou},
  title = {{MoCA}: Multi-modal Cross-masked Autoencoder for Time Series in Digital Health},
  howpublished = {arXiv:2506.02260}, year = {2025}, doi = {10.48550/arXiv.2506.02260}
}

@article{sasaki2016,
  author = {Jeffer Eidi Sasaki and Amanda M. Hickey and John W. Staudenmayer and Dinesh John and Jane A. Kent and Patty S. Freedson},
  title = {Performance of Activity Classification Algorithms in Free-Living Older Adults},
  journal = {Medicine \& Science in Sports \& Exercise}, volume = {48}, number = {5}, pages = {941--950}, year = {2016},
  doi = {10.1249/MSS.0000000000000844}
}

@misc{selfback2020dataset,
  author={{UCI Machine Learning Repository}},
  title={{selfBACK} [Dataset]},
  year={2020},
  doi={10.24432/C5SK65},
  url={https://doi.org/10.24432/C5SK65}
}

@article{yuan2024harnet,
  author = {Hang Yuan and Shing Chan and Andrew P. Creagh and Catherine Tong and Aidan Acquah and David A. Clifton and Aiden Doherty},
  title = {Self-supervised learning for human activity recognition using 700,000 person-days of wearable data},
  journal = {npj Digital Medicine}, volume = {7}, pages = {91}, year = {2024},
  doi = {10.1038/s41746-024-01062-3}
}

@article{chan2024capture24,
  author = {Shing Chan and Yuan Hang and Catherine Tong and Aidan Acquah and Abram Schonfeldt and Jonathan Gershuny and Aiden Doherty},
  title = {{CAPTURE-24}: A large dataset of wrist-worn activity tracker data collected in the wild for human activity recognition},
  journal = {Scientific Data}, volume = {11}, pages = {1135}, year = {2024},
  doi = {10.1038/s41597-024-03960-3}
}

@article{ganin2016,
  author = {Yaroslav Ganin and Evgeniya Ustinova and Hana Ajakan and Pascal Germain and Hugo Larochelle and Fran{\c{c}}ois Laviolette and Mario Marchand and Victor Lempitsky},
  title = {Domain-Adversarial Training of Neural Networks},
  journal = {Journal of Machine Learning Research}, volume = {17}, number = {59}, pages = {1--35}, year = {2016}
}

@article{courty2017,
  author = {Nicolas Courty and R{\'e}mi Flamary and Devis Tuia and Alain Rakotomamonjy},
  title = {Optimal Transport for Domain Adaptation},
  journal = {IEEE Transactions on Pattern Analysis and Machine Intelligence}, volume = {39}, number = {9}, pages = {1853--1865}, year = {2017},
  doi = {10.1109/TPAMI.2016.2615921}
}

@inproceedings{gupta2016,
  author = {Saurabh Gupta and Judy Hoffman and Jitendra Malik},
  title = {Cross Modal Distillation for Supervision Transfer},
  booktitle = {Proceedings of the IEEE Conference on Computer Vision and Pattern Recognition}, pages = {2827--2836}, year = {2016}
}

@inproceedings{elsayed2019,
  author = {Gamaleldin F. Elsayed and Ian Goodfellow and Jascha Sohl-Dickstein},
  title = {Adversarial Reprogramming of Neural Networks},
  booktitle = {International Conference on Learning Representations}, year = {2019}
}

@inproceedings{houlsby2019,
  author = {Neil Houlsby and Andrei Giurgiu and Stanislaw Jastrzebski and Bruna Morrone and Quentin de Laroussilhe and Andrea Gesmundo and Mona Attariyan and Sylvain Gelly},
  title = {Parameter-Efficient Transfer Learning for {NLP}},
  booktitle = {Proceedings of the 36th International Conference on Machine Learning}, pages = {2790--2799}, year = {2019}
}

@inproceedings{wang2021tent,
  author = {Dequan Wang and Evan Shelhamer and Shaoteng Liu and Bruno Olshausen and Trevor Darrell},
  title = {Tent: Fully Test-Time Adaptation by Entropy Minimization},
  booktitle = {International Conference on Learning Representations}, year = {2021}
}

@inproceedings{lipman2023,
  author = {Yaron Lipman and Ricky T. Q. Chen and Heli Ben-Hamu and Maximilian Nickel and Matt Le},
  title = {Flow Matching for Generative Modeling},
  booktitle = {International Conference on Learning Representations}, year = {2023}
}

@inproceedings{liu2023rectified,
  author = {Xingchao Liu and Chengyue Gong and Qiang Liu},
  title = {Flow Straight and Fast: Learning to Generate and Transfer Data with Rectified Flow},
  booktitle = {International Conference on Learning Representations}, year = {2023}
}

@inproceedings{liu2023i2sb,
  author = {Guan-Horng Liu and Arash Vahdat and De-An Huang and Evangelos A. Theodorou and Weili Nie and Anima Anandkumar},
  title = {{I$^2$SB}: Image-to-Image {Schr\"odinger} Bridge},
  booktitle = {Proceedings of the 40th International Conference on Machine Learning}, year = {2023}
}
\clearpage

\raggedbottom
\clearpage
\appendix
\section{PAAWS Data and Calibration Pairs}\label{app:pairing}
\paragraph{Signals.} Acceleration is resampled from 80 to 50\,Hz with a polyphase filter. Each nonoverlapping ten-second window contains $3\times500$ values in units of standard gravity. Lab and SimFL inputs use the device worn in the top position on each wrist. FL inputs use the available left and right wrist devices, whose top or bottom position is uncertain. All methods reverse the first axis for right-wrist windows and retain the released axes for left-wrist windows.

\paragraph{Thigh2Wrist pairs.} Each wrist window is paired with its concurrent right-thigh window. The 224,697 training pairs contain 113,091 unique thigh windows, each of which can accompany both wrists.

\paragraph{Lab2Freeliving pairs.} Independently timed SimFL and Lab windows are matched within participant--activity--wrist-side groups. Within each group, both lists are independently shuffled and paired by position, cycling the shorter list. The 1,226 groups yield 137,299 training pairs. Pairs are reshuffled each epoch, with the same schedule for all bridge variants within a run.

\paragraph{Feature standardization.} Coordinatewise means and population standard deviations are computed without pair weights over 132,279 Lab wrist windows of training participants for Lab2Freeliving and 113,091 unique thigh windows of training participants for Thigh2Wrist. Standard deviations are floored at $10^{-4}$. Statistics remain fixed during fitting and evaluation.

\section{Models and Optimization}\label{app:architecture}
\paragraph{Reference models.} Frozen MoCA produces 768-dimensional features. In Thigh2Wrist, a linear classifier reads these features, so $d=768$. In Lab2Freeliving, a projection block comprising a linear $768\rightarrow128$ layer, layer normalization, and GELU is trained with a linear four-class classifier. Both are then frozen. The bridge acts on the standardized output of this block, so $d=128$, and the classifier reads the inverse-standardized bridge output.

The three additional encoders produce 128-dimensional features, so $d=128$. DeepConvLSTM uses four 64-channel temporal convolutions with kernel size 5 and two 128-unit LSTM layers with dropout 0.1. InceptionTime uses six modules with 32-channel branches, kernels 41, 21, and 11, and residual connections every three modules. The accelerometer-only HART uses stride-20 patches, 25 tokens of dimension 128, and four blocks combining four-head attention with a depthwise convolution of kernel size 5 and a 256-unit feed-forward layer.

\paragraph{Bridge architectures.} All bridges use two SiLU hidden layers, with width 768 in MoCA Thigh2Wrist on PAAWS and selfBACK, and 256 elsewhere. Mapper and Distillation use residual multilayer perceptrons. MeanFlow receives the path point $z(t)$, the wrist feature $z_s$, and three 32-dimensional sinusoidal embeddings of $t_0$, $t$, and $t-t_0$. Output layers are zero-initialized. The encoder, classifier, and standardization remain frozen.

\paragraph{MeanFlow time sampling.}\label{app:meanflow-sampling}
For each pair, we apply a sigmoid to two independent $\mathcal{N}(-0.4,1^2)$ draws and sort them into $t_0\leq t$. For a uniformly selected half of each batch, we set $t_0=t$ at the sampled upper time. The loss stabilizer is $\epsilon=0.01$ in Lab2Freeliving and $0.06$ in primary MoCA Thigh2Wrist.

\paragraph{Lab2Freeliving training.} Reference class weights are $132{,}279/(4N_k)$, where $N_k$ counts Lab training windows of activity $k$. Each bridge pair in participant--activity--wrist-side group $g$ receives weight $137{,}299/(8G_gn_g)$. Here $n_g$ is the larger of the group's SimFL and Lab counts, and $G_g$ counts the participant groups that share the activity and wrist side of $g$. Training averages the weighted per-pair losses in each batch. Additional encoders train jointly with their classifiers on Lab data and are frozen before bridge fitting. Table~\ref{tab:training} gives optimization settings.

\begin{table}[htbp]
\setlength{\belowcaptionskip}{6pt}
\centering
\caption{Lab2Freeliving optimization settings. All methods use AdamW and batches of 128, and bridges clip gradient norms at 1.}
\label{tab:training}
\small
\begin{tabular}{lccc}
\toprule
Setting & MoCA projection/classifier & Other encoders & Bridges\\
\midrule
Learning rate & $10^{-4}$ & $10^{-3}$ & $10^{-4}$\\
Weight decay & $0.05$ & $10^{-4}$ & $10^{-4}$\\
Epochs & 30 & 30 & 5--30\\
Precision & Mixed & Mixed & Full\\
\bottomrule
\end{tabular}
\end{table}

\paragraph{Primary Thigh2Wrist training.}\label{app:wrist-training}
The thigh classifier uses weights that equalize session--activity groups, where the session is Lab or SimFL. AdamW uses learning rate $10^{-4}$ and weight decay 0.05. Each bridge-training pair in session--activity--wrist-side group $g$ receives weight $224{,}697/(16n_g)$, where $n_g$ is the group count. Bridge learning rate and weight decay are $10^{-4}$, with gradient clipping at 1. Reference and bridge training use batches of 128, full float32 precision, and 30 epochs.

\paragraph{Matched comparisons.} PAAWS comparisons (Tables~\ref{tab:wrist-controls} and~\ref{tab:joint-finetuning}) share the primary Thigh2Wrist data, weights, optimizer, and 30-epoch schedule. The Direct wrist classifier and joint fine-tuning both initialize from the thigh classifier. The Direct wrist classifier updates the classifier alone, and joint fine-tuning also updates the encoder. The activity-only bridge trains the MeanFlow architecture with the cross-entropy term alone. These comparisons use seeds 20260717, 20260718, and 20260719. The activity-only results use three fits from the same training implementation. The MeanFlow row summarizes the same three primary runs as Table~\ref{tab:main}.

\paragraph{Additional Thigh2Wrist encoders.}\label{app:additional-backbones}
DeepConvLSTM, InceptionTime, and HART train jointly with linear thigh classifiers from scratch on the 113,091 unique training-thigh windows, then remain frozen. Reference fitting uses AdamW with learning rate $10^{-3}$, weight decay $10^{-4}$, batches of 128, and 30 epochs. Bridge fitting follows the primary Thigh2Wrist protocol with hidden width 256 and MeanFlow stabilizer $\epsilon=0.01$. All training uses float32 precision.

\section{Evaluation and Uncertainty}\label{app:evaluation}
\paragraph{Checkpoint selection.} Lab2Freeliving selects the earliest maximum macro-F1 on 33,502 Lab-validation windows for the reference and 22,742 SimFL-validation windows for the bridge. Bridges train for 5--30 epochs with early-stopping patience of five epochs and a minimum improvement of $10^{-4}$. All Thigh2Wrist comparisons use the final checkpoint after 30 epochs. Checkpoints are fixed before FL scoring.

\paragraph{Metrics and class support.} Metrics pool windows across held-out participants. Measured-thigh scores use the same wrist-pair weighting as wrist predictions. Table~\ref{tab:counts} gives class support for both four-class PAAWS designs.

\begin{table}[htbp]
\setlength{\belowcaptionskip}{6pt}
\centering
\caption{PAAWS four-class test wrist windows. Every method and run within a calibration design uses the same windows. Thigh2Wrist requires a concurrent thigh recording.}
\label{tab:counts}
\begin{tabular}{lrr}
\toprule
Activity & Thigh2Wrist & Lab2Freeliving\\
\midrule
Cycling & 9,552 & 9,552\\
Sitting & 1,261,789 & 1,267,245\\
Standing & 176,233 & 177,148\\
Walking & 140,795 & 141,826\\
\midrule
Total & 1,588,369 & 1,595,771\\
\bottomrule
\end{tabular}
\end{table}

\paragraph{Five-window averaging.} Five-window averaging combines class probabilities from the current and up to four preceding windows. History uses consecutive ten-second windows within each participant and wrist stream and resets at recording gaps. Averaging precedes the activity scoring filter in both the four-class and six-class analyses. Gains are computed within each run before averaging.

\paragraph{Participant-bootstrap intervals.} For each of 5,000 bootstrap draws, we sample participants with replacement and retain all their windows. The same resample is applied across methods and runs. We recompute pooled macro-F1 within each run. Paired differences average the three runs. The 95\% participant-bootstrap intervals are pointwise percentile intervals and describe participant variation conditional on the fitted models. Run-to-run sample standard deviation is reported separately.

\begin{table}[htbp]
\setlength{\belowcaptionskip}{6pt}
\centering
\caption{Macro-F1 differences, MeanFlow minus comparator, in percentage points, with paired 95\% participant-bootstrap intervals.}
\label{tab:bootstrap}
\small
\begin{tabular}{llrr}
\toprule
Data and design & Comparator & Difference & 95\% interval\\
\midrule
PAAWS Thigh2Wrist & Unadapted reference & 20.91 & 17.31 to 24.61\\
PAAWS Lab2Freeliving & Unadapted reference & 9.54 & 1.74 to 13.52\\
selfBACK Thigh2Wrist & Unadapted reference & 41.01 & 31.70 to 50.59\\
CAPTURE-24 transfer & Unadapted reference & 11.98 & 8.33 to 14.60\\
\midrule
selfBACK Thigh2Wrist & Direct wrist classifier & 3.16 & $-2.10$ to 7.89\\
CAPTURE-24 transfer & Direct wrist classifier & $-0.85$ & $-3.90$ to 2.31\\
PAAWS Thigh2Wrist & Direct wrist classifier & 1.38 & 0.59 to 2.08\\
PAAWS Thigh2Wrist & Activity-only bridge & 2.77 & 1.81 to 4.48\\
\bottomrule
\end{tabular}
\end{table}

\paragraph{Feature visualization.}\label{app:visualization}
For Thigh2Wrist (Figure~\ref{fig:thigh2wrist-umap}), UMAP is fitted on 10,712 unique SimFL-validation thigh features. The plot displays 128 validation references and computes class means over 8,192 held-out wrist--thigh pairs. For Lab2Freeliving (Figure~\ref{fig:umap}), UMAP is fitted on 128 Lab-validation references, and class means use the 2,048 displayed wrist recordings. Each view uses one run.

\begingroup
\setlength{\intextsep}{6pt}
\section{Class-Level Errors on PAAWS}\label{app:additional}
Cells in Figures~\ref{fig:confusion-thigh} and~\ref{fig:confusion} show mean row percentages over three runs on the same test windows within each comparison. Rows are true activities and columns are current-window predictions.
\begin{figure}[!htbp]
\centering
\includegraphics[width=\linewidth]{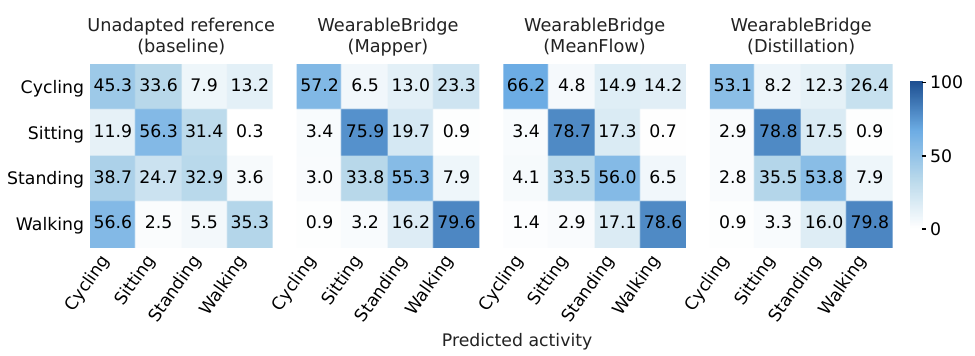}
\caption{PAAWS Thigh2Wrist with MoCA on the same 1,588,369 wrist windows from 40 participants. Methods are defined as in Table~\ref{tab:main}, and Table~\ref{tab:counts} gives class counts.}
\label{fig:confusion-thigh}
\end{figure}

\begin{figure}[!htbp]
\centering
\includegraphics[width=\linewidth]{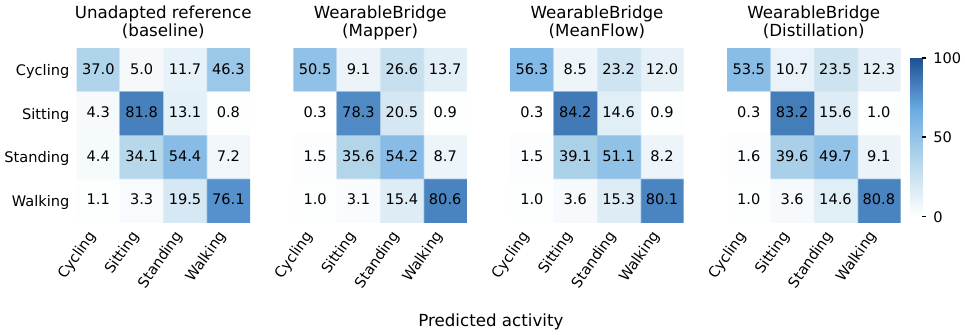}
\caption{PAAWS Lab2Freeliving with MoCA on the same 1,595,771 wrist windows from 40 participants. Methods are defined as in Table~\ref{tab:main}, and Table~\ref{tab:counts} gives class counts.}
\label{fig:confusion}
\end{figure}

\endgroup

\appendixfigurebreak
\section{selfBACK Paired Replication}
\label{app:selfback-setup}
\paragraph{Data and pairing.} selfBACK supplies simultaneous wrist--thigh recordings for nine activities: lying, sitting, standing, jogging, ascending and descending stairs, and walking at three speeds. The participant split is 20/6/7 for training/validation/test. Streams are aligned on absolute timestamps, interpolated onto a 100\,Hz grid using gaps of at most 30\,ms, and resampled into complete ten-second windows at 50\,Hz. This gives 2,828/874/963 paired windows. Wrist handedness is unspecified in the released metadata, so inputs retain the released axes.

\paragraph{Training and evaluation.} Each run fits a nine-class thigh classifier on frozen 768-dimensional MoCA features, then freezes it for bridge training. The Direct wrist classifier initializes from the same thigh classifier. All methods share training windows and class-balanced weights. Bridge standardization uses the 2,828 training-thigh windows. Architectures, losses, and optimizer settings follow primary Thigh2Wrist (Appendix~\ref{app:wrist-training}), with gradient clipping at 1 for all fitted models and final checkpoints after 30 epochs. The six validation participants are reserved and unused in fitting or checkpoint selection. Current-window evaluation uses three runs and the participant-bootstrap procedure in Appendix~\ref{app:evaluation}.

\section{CAPTURE-24 Transfer Without Target Labels}
\label{app:capture-sourceonly}
\paragraph{Data and preprocessing.} We evaluate 134,658 windows from 50 CAPTURE-24 participants, P102--P151 \citep{chan2024capture24}. Fully annotated ten-second windows receive their modal label and are restricted to cycling, sitting, standing, and walking, with bicycling mapped to cycling. After 50\,Hz preprocessing, signals are interpolated to 30\,Hz for frozen HARNet10 features (1,024 dimensions) pretrained on UK Biobank \citep{yuan2024harnet}. CAPTURE-24 devices are worn on the dominant wrist, and all methods treat them as right wrists.

\paragraph{Training on PAAWS and evaluation.} Fitting and standardization use PAAWS training participants, and checkpoint selection uses PAAWS validation participants. CAPTURE-24 labels define the evaluation set and metrics. The Lab reference is a linear $1{,}024\rightarrow4$ classifier trained with AdamW at learning rate $10^{-3}$ and weight decay $10^{-4}$ for 30 epochs, using batches of 128 and float32 precision. The Direct wrist classifier initializes from this Lab classifier and updates it on labeled PAAWS SimFL features while HARNet10 stays frozen. MeanFlow keeps both the encoder and the Lab classifier fixed and learns the bridge. Both share SimFL examples, weights, optimizer, maximum budget, and selection rule, following Lab2Freeliving (Appendices~\ref{app:architecture} and~\ref{app:evaluation}) with method-specific early stopping. Evaluation uses current-window predictions, three runs, and the participant-bootstrap procedure in Appendix~\ref{app:evaluation}. Table~\ref{tab:capture-sourceonly} gives overall and class-level scores.

\begin{table}[!htbp]
\setlength{\belowcaptionskip}{6pt}
\centering
\caption{CAPTURE-24 transfer in the Lab2Freeliving design with frozen HARNet10, in percent (mean $\pm$ sample SD over three runs), on 134,658 windows from 50 participants (2,409 cycling, 100,733 sitting, 9,592 standing, and 21,924 walking). Fitting and selection use PAAWS only. The unadapted reference is the baseline: it applies the frozen PAAWS Lab classifier to CAPTURE-24 wrist features. The Direct wrist classifier updates that classifier on labeled PAAWS SimFL windows.}
\label{tab:capture-sourceonly}
\footnotesize
\setlength{\tabcolsep}{1.2pt}
\begin{tabular*}{\linewidth}{@{\extracolsep{\fill}}lccccccc@{}}
\toprule
 & \multicolumn{3}{c}{Overall} & \multicolumn{4}{c}{Class F1}\\
\cmidrule(lr){2-4}\cmidrule(l){5-8}
Method & Macro-F1 & Bal.\ acc. & Accuracy & Cycling & Sitting & Standing & Walking\\
\midrule
Unadapted reference {\scriptsize(baseline)} & 42.75\sd{0.01} & 47.09\sd{0.04} & 68.78\sd{0.06} & 5.38\sd{0.14} & 84.12\sd{0.05} & 26.23\sd{0.04} & 55.27\sd{0.05}\\
\midrule
MeanFlow & 54.73\sd{1.58} & 57.50\sd{1.18} & 69.40\sd{0.94} & 50.64\sd{4.94} & 83.16\sd{0.84} & 22.57\sd{0.45} & 62.56\sd{0.63}\\
\midrule
Direct wrist classifier & 55.58\sd{0.42} & 65.17\sd{0.54} & 72.17\sd{0.48} & 50.30\sd{2.63} & 85.81\sd{0.39} & 25.15\sd{0.14} & 61.08\sd{0.83}\\
\bottomrule
\end{tabular*}
\end{table}

\Needspace{28\baselineskip}
\section{Additional Encoders, Temporal Context, and Fine-Tuning}
\label{app:secondary-metrics}
Table~\ref{tab:backbones} reports macro-F1 for the additional reference encoders, and Table~\ref{tab:temporal} reports macro-F1 with five-window averaging for the primary MoCA references. Table~\ref{tab:joint-finetuning} gives the separate encoder-adaptation comparison.

\begin{table}[!htbp]
\setlength{\belowcaptionskip}{6pt}
\centering
\caption{Additional reference encoders: current-window macro-F1 (\%), mean $\pm$ sample SD over three runs. The unadapted reference is the baseline: it applies each encoder's frozen reference classifier to wrist features. Mapper, MeanFlow, and Distillation map wrist features before that classifier.}
\label{tab:backbones}\label{tab:wrist-thigh-backbones}
\footnotesize
\setlength{\tabcolsep}{2.5pt}
{\raggedright\small\textbf{(a) Thigh2Wrist.} 1,588,369 wrist windows from 40 participants.\par}
\smallskip
\begin{tabular*}{\linewidth}{@{\extracolsep{\fill}}lcccc@{}}
\toprule
 & Baseline & \multicolumn{3}{c}{WearableBridge}\\
\cmidrule(lr){2-2}\cmidrule(l){3-5}
Reference encoder & Unadapted reference & Mapper & MeanFlow & Distillation\\
\midrule
DeepConvLSTM & 33.73\sd{4.30} & 43.93\sd{0.93} & 43.36\sd{1.22} & 44.20\sd{0.92}\\
InceptionTime & 32.49\sd{5.47} & 52.04\sd{1.71} & 52.37\sd{2.66} & 52.63\sd{1.38}\\
HART & 31.67\sd{2.78} & 48.21\sd{0.73} & 49.20\sd{0.72} & 47.90\sd{0.59}\\
\bottomrule
\end{tabular*}\par
\medskip
{\raggedright\small\textbf{(b) Lab2Freeliving.} 1,595,771 wrist windows from the same 40 participants.\par}
\smallskip
\begin{tabular*}{\linewidth}{@{\extracolsep{\fill}}lcccc@{}}
\toprule
 & Baseline & \multicolumn{3}{c}{WearableBridge}\\
\cmidrule(lr){2-2}\cmidrule(l){3-5}
Reference encoder & Unadapted reference & Mapper & MeanFlow & Distillation\\
\midrule
DeepConvLSTM & 57.10\sd{0.96} & 69.47\sd{0.41} & 69.76\sd{0.26} & 69.15\sd{1.02}\\
InceptionTime & 60.65\sd{0.24} & 68.82\sd{1.76} & 68.78\sd{0.76} & 68.97\sd{0.64}\\
HART & 56.37\sd{0.28} & 68.91\sd{0.33} & 68.98\sd{0.50} & 69.14\sd{0.81}\\
\bottomrule
\end{tabular*}
\end{table}

\begin{table}[!htbp]
\setlength{\belowcaptionskip}{6pt}
\centering
\caption{Macro-F1 with five-window averaging, mean $\pm$ sample SD over three runs. $\Delta$ is the within-run gain over the same method in Table~\ref{tab:main}, in percentage points (pp). Methods are defined as in Table~\ref{tab:main}.}
\label{tab:temporal}
\footnotesize
\setlength{\tabcolsep}{2.5pt}
\begin{tabular*}{\linewidth}{@{\extracolsep{\fill}}lcccc@{}}
\toprule
 & \multicolumn{2}{c}{Thigh2Wrist} & \multicolumn{2}{c}{Lab2Freeliving}\\
\cmidrule(lr){2-3}\cmidrule(l){4-5}
Method & Macro-F1 (\%) & $\Delta$ macro-F1 (pp) & Macro-F1 (\%) & $\Delta$ macro-F1 (pp)\\
\midrule
Unadapted reference {\scriptsize(baseline)} & 35.53\sd{0.25} & +0.46 & 58.16\sd{1.70} & +4.16\\
Mapper & 60.28\sd{0.78} & +6.15 & 66.49\sd{0.24} & +5.59\\
MeanFlow & 61.70\sd{0.75} & +5.72 & 68.90\sd{1.42} & +5.36\\
Distillation & 60.93\sd{0.43} & +5.87 & 67.55\sd{1.32} & +5.26\\
\bottomrule
\end{tabular*}
\end{table}

\paragraph{Joint encoder fine-tuning.}\label{app:joint-finetuning}
Joint fine-tuning updates both the MoCA encoder and the classifier, initialized from the thigh classifier, on wrist data. Table~\ref{tab:joint-finetuning} uses the same three seeds as Table~\ref{tab:wrist-controls}, with the training protocol in Appendix~\ref{app:architecture}. Joint fine-tuning trains 85.13 million encoder parameters in addition to the classifier. MeanFlow trains a bridge with 2.44 million parameters and keeps the reference encoder and classifier fixed. The fine-tuning result provides context for adaptation of the full reference model.

\begin{table}[!htbp]
\setlength{\belowcaptionskip}{6pt}
\centering
\caption{PAAWS Thigh2Wrist with a frozen encoder or joint encoder fine-tuning. Current-window scores are percentages (mean $\pm$ sample SD over three runs) on the same FL test windows as Table~\ref{tab:main}.}
\label{tab:joint-finetuning}
\footnotesize
\setlength{\tabcolsep}{2.5pt}
\begin{tabular*}{\linewidth}{@{\extracolsep{\fill}}lccc@{}}
\toprule
Method & Macro-F1 & Bal.\ acc. & Accuracy\\
\midrule
MeanFlow (frozen encoder) & 55.98\sd{0.51} & 69.87\sd{0.73} & 76.11\sd{1.64}\\
Joint fine-tuning & 62.56\sd{0.59} & 72.82\sd{1.88} & 79.62\sd{0.86}\\
\bottomrule
\end{tabular*}
\end{table}

\Needspace{28\baselineskip}
\section{Six-Class Thigh2Wrist Extension}\label{app:six-class}
\paragraph{Calibration and fitting.} We add lying and stairs and retain the 170/43 training/validation participant split. Concurrent Lab and SimFL wrist--thigh recordings provide 250,084 training pairs and 59,350 validation pairs, with lying supplied by Lab. Frozen MoCA features support a six-class linear thigh classifier. Reference fitting and unweighted bridge standardization use 125,865 unique training-thigh windows. Reference weights equalize activities and then observed sessions within each activity. Adaptation weights equalize activities and then observed session--wrist-side cells within each activity. The Direct wrist classifier and MeanFlow share calibration rows, activity weights, and the initial thigh classifier. Architectures and optimization follow primary Thigh2Wrist (Appendix~\ref{app:wrist-training}), with final checkpoints after 30 epochs and three runs.

\paragraph{Evaluation and results.} We evaluate the same 40 held-out FL participants on 1,835,395 eligible wrist windows with concurrent right-thigh recordings. Macro-F1 averages all six activities on this expanded population. Intervals use the 5,000 paired participant-bootstrap draws described in Appendix~\ref{app:evaluation}. Relative to the Direct wrist classifier, MeanFlow improves current-window macro-F1 for 39 of 40 participants and F1 for all six activities, averaged over three runs (Table~\ref{tab:six-class}).

\begin{table}[!htbp]
\setlength{\belowcaptionskip}{6pt}
\centering
\caption{Six-class Thigh2Wrist results with frozen MoCA and wrist-only inference. (a) Overall macro-F1 in percent, mean $\pm$ sample SD over three runs. Parentheses contain paired 95\% participant-bootstrap intervals for MeanFlow minus the Direct wrist classifier, in percentage points (pp). (b) Current-window class F1, averaged over three runs, and test wrist-window counts.}
\label{tab:six-class}
\small
\setlength{\tabcolsep}{4pt}
{\raggedright\textbf{(a) Overall macro-F1.}\par}
\smallskip
\begin{tabular*}{\linewidth}{@{\extracolsep{\fill}}lc@{}}
\toprule
Method & Macro-F1\\
\midrule
Unadapted reference & 14.46\sd{0.12}\\
Direct wrist classifier & 40.55\sd{0.34}\\
MeanFlow & 44.04\sd{0.69}\\
\midrule
MeanFlow minus Direct (pp) & +3.49 (2.58 to 4.34)\\
\bottomrule
\end{tabular*}\par
\medskip
{\raggedright\textbf{(b) Class F1 and support.}\par}
\smallskip
\begin{tabular*}{\linewidth}{@{\extracolsep{\fill}}lrrr@{}}
\toprule
Activity & Test windows & Direct wrist classifier & MeanFlow\\
\midrule
Cycling & 9,552 & 13.82 & 18.66\\
Lying & 241,633 & 43.37 & 44.47\\
Sitting & 1,261,789 & 67.75 & 70.40\\
Stairs & 5,393 & 9.88 & 16.25\\
Standing & 176,233 & 37.47 & 38.22\\
Walking & 140,795 & 71.02 & 76.22\\
\bottomrule
\end{tabular*}
\end{table}

\end{document}